\documentclass[journal,twoside]{{IEEEtran}}
\usepackage{amsfonts}
\usepackage{amssymb}
\usepackage{graphicx}
\usepackage{caption}
\usepackage{subcaption}
\usepackage{color}
\usepackage[ruled,vlined]{algorithm2e}

\begin{document}
\title{HIN-RNN: A Graph Representation Learning Neural Network for Fraudster Group Detection With No Handcrafted Features
}
%
%
\author{
Saeedreza Shehnepoor*,
Roberto Togneri,
Wei Liu,
Mohammed Bennamoun,
\thanks{S. Shehnepoor (*corresponding author) is with the University of Western Australia, Perth, Australia.
R. Togneri is with the University of Western Australia, Perth, Australia.
M. Bennamoun is with the University of Western Australia, Perth, Australia.
W. Liu is with the University of Western Australia, Perth, Australia.
emails: \{saeedreza.shehnepoor@research.uwa.edu.au, roberto.togneri@uwa.edu.au, wei.liu@uwa.edu.au, mohammed.bennamoun@uwa.edu.au.\}}
}
%
%
%
\maketitle              
\begin{abstract}
Social reviews are indispensable resources for modern consumers' decision making. For financial gain, companies pay fraudsters preferably in groups to demote or promote products and services since consumers are more likely to be misled by a large amount of similar reviews from groups. Recent approaches on fraudster group detection employed handcrafted features of group behaviors without considering the semantic relation between reviews from the reviewers in a group. In this paper, we propose the first neural approach, HIN-RNN, a Heterogeneous Information Network (HIN) Compatible RNN for fraudster group detection that requires no handcrafted features. HIN-RNN provides a unifying architecture for representation learning  of each reviewer, with the initial vector as the sum of word embeddings of all review text written by the same reviewer, concatenated by the ratio of negative reviews. Given a co-review network representing reviewers who have reviewed the same items with the same ratings and the reviewers' vector representation, a collaboration matrix is acquired through HIN-RNN training. The proposed approach is confirmed to be effective with marked improvement over state-of-the-art approaches on both the Yelp (22\% and 12\% in terms of recall and F1-value, respectively) and Amazon (4\% and 2\% in terms of recall and F1-value, respectively) datasets. 
\end{abstract}

\section{Introduction}
\label{sec:intro}
Modern consumers trust more the reviews from other fellow consumers who used the products or services than advertisements. For financial gain, fabricated fraud reviews are rife on social review platforms with the purpose of either promoting one's own products or to demote rivals’ products or both. Even worse, fraudsters may form groups that collectively attack (or promote) products
for different purposes such as dominance over the sentiment of a target product, or sharing the overall workload of the fraud review effort~\cite{Mukherjee2012,Lingyun2018,MukherjeeV0G13,Ye2015}.
Since the first proposed approach on fraud detection~\cite{Jindal2008}, most studies focused on detecting individual fraudsters~\cite{MukherjeeV0G13,shehne2017}. Only recently has there been an increased research effort targeting fraudster groups~\cite{Allah2013,JI2020454,Zhang2020,Xu2019}. Fraudster groups are more effective in misleading consumers due to the volume and consistency of their coordinated group reviews. Consequently, they do more damage to the reputation of a review platform by undermining the trust of consumers.
Fraudster groups detection is a much more challenging task than individual fraudster detection because, fraudster groups can distribute their suspicious behavior over all fraudsters in the group, and through camouflage so no single fraudster stands out~\cite{Mukherjee2012,Ye2015}. These groups can also produce batches of fake reviews more severely in a short span of time, affecting the reputation of the target product. \\
Recent approaches for fraudster group detection are dominated by Frequent Itemset Mining (FIM) based algorithms~\cite{Allah2013,Xu2019}, and graph based approaches~\cite{JI2020454,Zhang2020}. FIM-based methods, generally utilize a two-step approach to first find the candidate groups and then rank them, 
whereas graph-based approaches rely on graph partition or clustering algorithms~\cite{Ye2015} for determining candidates. However, both suffer from various limitations as outlined below. \\
\textbf{First}, approaches such as 
Ji \textit{et al.}~\cite{JI2020454}, and Zhang \textit{et al.} ~\cite{Zhang2020} use handcrafted group-level features (e.g., group size, or group rating deviation) that are generally unable to capture real behaviors of many reviewers in a group. 
\textbf{Second},
members in a group not only perform collective activities, but also aim to increase the impression by maintaining consistency in their review semantics~\cite{Lingyun2018}.
The semantic dependencies between reviewers is demonstrated to play an important role in 
fraudster (single user or fraudster groups) activities~\cite{Shebuit2015,shehne2017}, and were overlooked in fraudster group detection. 
\textbf{Third}, a group of reviewers could form a genuine group who write reviews on different items out of the same interest. This is likely to happen when different reviewers write reviews on similar items (i.e. \textit{co-reviewing}), often with multiple interaction histories. 
A fraudster in a genuine group (i.e. a \textbf{fraudster imposter}) could lead to the group being detected as a fraudster group using the aforementioned approaches. In such cases, the genuine groups, will be wrongly predicted as fraud. This leads to a higher False Positive (FP) rate. Conversely, fraudster groups could camouflage themselves by writing reviews that are similar to the ones written by genuine reviews, to avoid detection. This increases the False Negative (FN) rate of the classification.\\ 
To address these challenges, we propose a novel framework with four steps: \textbf{first}, we employ the Sum of Word Embeddings (SoWEs) inspired by the theory of Collective Intelligence (CI)~\cite{Malone2010,malone2009} to use all reviews written by a reviewer as features to overcome the limitations of handcrafted features. 
Sum of Word Embeddings (SoWEs) have been verified in various occasions to surpass more complex document embeddings~\cite{Zhang2016} as a simple yet effective representation for documents. 
In this research, we therefore choose to use SoWE of all tokens in a review as a review's vector representation. 
A reviewer’s vector representation is thus the SoWE of all reviews written by this reviewer and a group’s vector representation is thus the average of vector representations of all reviewers in that group. The SoWE is further fine-tuned by training a Convolutional Neural Network (CNN). A CNN is chosen in preference to Recurrent Neural Networks (RNNs) because CNNs are better in dealing with the potential multiple aspects covered in each review~\cite{khan2018,Ren:2017:NND:3043977.3044107}. \textbf{Next}, we determine the candidate groups based on the time interval between written reviews. Fraudsters in a group complete their reviews in a relatively shorter time window as compared with genuine reviewers. Previous studies estimated the time-interval to be 28 days~\cite{Mukherjee2012,JI2020454,Mukherjee2013}. We use the same time-interval (i.e. 28 days) to determine the possible collaboration between reviewers. Hence, every two reviewers co-reviewing 
at least two items with the same rating, form a connection (not necessarily in terms of fraudster activity). The outputs of this step are subgraphs 
that capture possible collusion between reviewers. \textbf{In the third step}, we propose ``HIN-RNN", i.e., Heterogeneous Information Network (HIN) compatible RNN. The HIN-RNN model takes the vector representations of reviewers from the first step and the resulting subgraphs from the second step to encode the non-local semantic dependencies (long range relationships) between reviewers through an autoregressive model. The HIN-RNN model
addresses the limitations of another graph generation model, i.e. GraphRNN, which generates subgraphs with only nodes of the same type. The output of this step is a collaboration matrix indicating the collaboration between the reviewers. 
\textbf{Finally,} given the collaboration matrix of groups, 
we exclude those reviewers with minimum connections, which are genuine reviewers unintentionally contributing to a fraud activity (thereby reducing FP), or fraudsters trying to camouflage to escape detection (thereby reducing FN). Then an average of the remaining reviewers' representations is fed to a simple fully connected layer to label (genuine/fraudster) each group. 
We can summarize our contributions as follows:
\begin{itemize}
    \item 
    We propose a novel approach called HIN-RNN that extends RNN to support the generation of different node types in a graph or network. 
    The proposed approach showed an average increase of 22\%, and 12\% for recall and F1-value, respectively, over most recent methods on the Yelp dataset (See Sec. \ref{sec:comparison}). Note that although the performance of the HIN-RNN is measured on the fraudster group detection task with only two types of reviewers involved (fraudster and genuine), such an approach can be applied to a broad range of graph generation tasks with multiple node types in the graph (See Sec. \ref{sec:representation-effect}).
    \item For the first time, we employ the semantic/behavioral representations (SoWE) of the reviewers to represent groups. Employing SoWE shows a boost in performance by 12\%, and 7\% for recall and F1-value on Yelp, respectively, as compared with the handcrafted features employed by Ji \textit{et al.}~\cite{JI2020454} (See Sec. \ref{sec:reviewer-rep-vs-group-features}). 
    \item We use the collaboration matrix to exclude reviewers with minimum connections (deviant reviewers) from a group. The exclusion of such deviant reviewers improves the performance of the proposed approach on Yelp (Amazon) substantially by 7\% (8\%), 12\% (7\%), 9\% (7\%) in terms of precision, recall, and F1-value, respectively (See Sec. \ref{sec:reviewer-removal-effect}). 
     
\end{itemize}
The rest of the paper is structured as follows. In Section \ref{sec:related-works}, we present the related work. In Section \ref{sec:proposed-approach}, we introduce our methodology. In Section \ref{sec:experimental-evaluation}, we provide the experimental evaluation. We conclude the paper with an outlook to future work in Section \ref{sec:conclusion}.

\section{Related Works}
\label{sec:related-works}
Studies on fraudster group detection fall under two groups, Frequent Itemset Mining (FIM) or Graph-based approaches.\\
\subsubsection{Frequent Itemset Mining} Frequent Itemset Mining (FIM) refers to approaches that basically assume members with the same set of reviewed items (itemset) as a candidate group~\cite{Agrawal94}. Such an assumption can also be extended to other aspects of reviewers' relations in different studies. Allahbakhsh \textit{et al.}~\cite{Allah2013} propose a data model where two types of reviewers and items connect via rating activities. Fraudster groups are detected by a biclique detector. A biclique refers to a group of reviewers who not only write reviews (FIM basic assumption), but also rate similarly the same group of items. 
After determining the candidate groups using the biclique criteria, different handcrafted fraudster group activity indicators (i.g. Group Rating Value Similarity (GVS), Group Rating Time Similarity (GTS)) 
are used to measure the potential of a group being fraudulent. A scoring function was applied to integrate these group features to obtain the final scores.
Xu \textit{et al.}~\cite{Xu2019} 
chose FIM to generate the candidate groups as at least two reviewers with at least three co-reviewed businesses. A homogeneity-based-Collusive Behavior Measure (h-CBM) are then proposed based on the targeted item, rating, temporal traits, and reviewer activity. The groups are scored through an unsupervised scoring model called Latent Collusive Model (LCM). In a nutshell, FIM approaches come with different shortcomings such as overlooking the deviant reviewers (i.e. single genuine reviewer in fraudster group and vice-versa) and the importance of the semantic connection between reviewers in a group~\cite{Allah2013,Xu2019}.\\
\subsubsection{Graph-based} 
\begin{figure}[h]
\centering
\includegraphics[width=\linewidth]{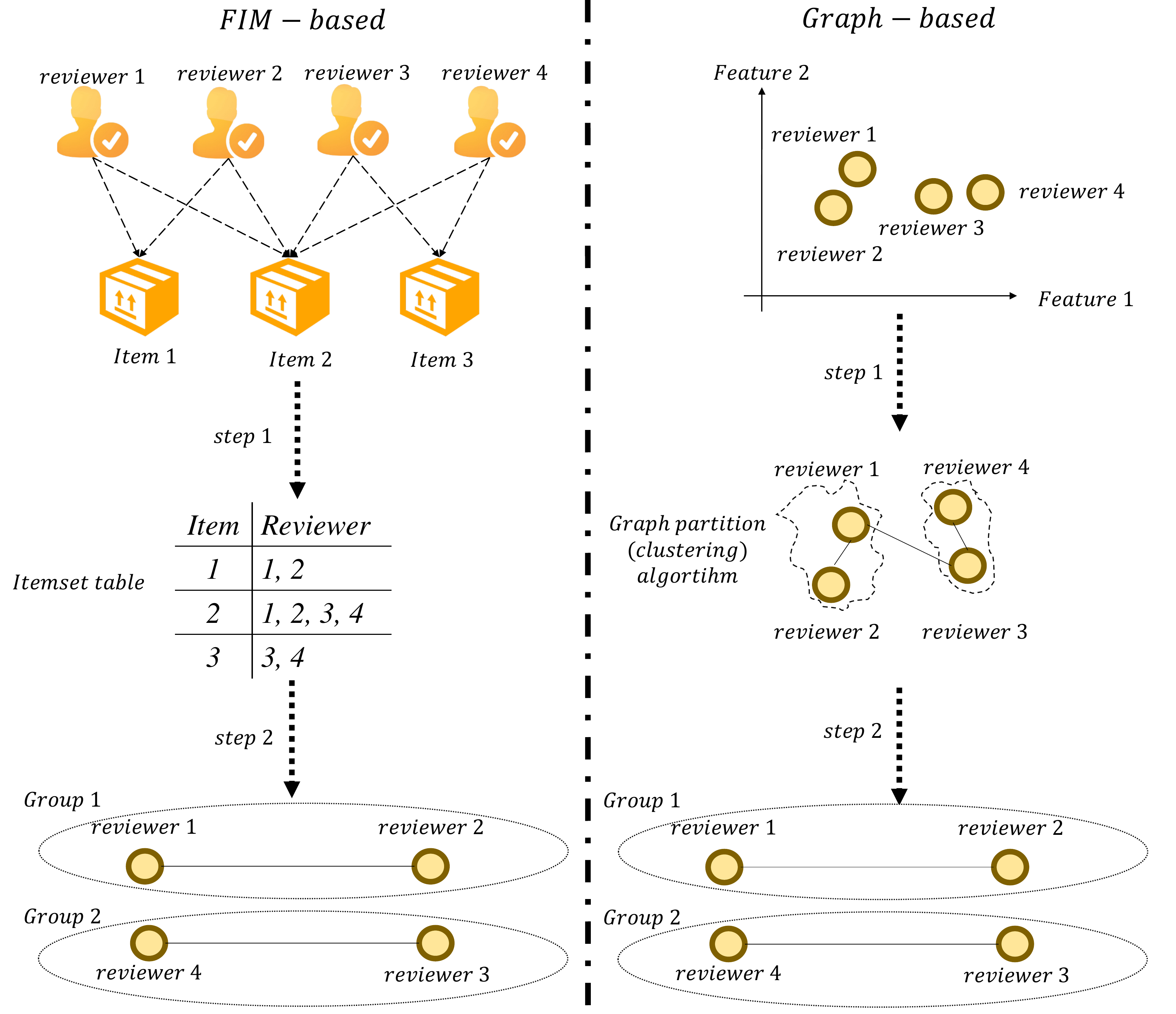}
\caption{A comparison of FIM-based and graph-based approaches.} 
\label{fig:fim-vs-graph}
\end{figure}

Graph-based fraudster group detection utilizes graph-based methods such as graph partition or clustering algorithms to address FIM model's limitations. 
Ji \textit{et al.}~\cite{JI2020454} propose a method where products are the focal points to overcome the limitations of FIM's reviewer-focused approaches. First, eight handcrafted group-level features (Group Rating Deviation, Group Size, Group Review Tightness, Group One Day Reviews, Group Extreme Rating Ratio, Group Co-Activeness, and Group Co-Active Review Ratio), six individual fraudster features (Ratio of Extreme Rating, Rating Deviation, The most Reviews One-day, Review Time Interval, Account Duration, and Active Time Interval Reviews), and three product related features (Product Rating Distribution, Product Average Rating Distribution, and Suspicious Score) are extracted.
Although such handcrafted features have been broadly employed for the fraud detection task, they could easily be manipulated by fraudsters or fraudster groups to fool a detection system. For example, fraudster groups can be separated into smaller groups to avoid detection~\cite{Zhang2020}. Some features can also be misleading. For example, the Ratio of Extreme Rating, and Rating Deviation, can result in different indications; if the Ratio of Extreme Rating increases for a group, the deviation is decreased which neutralizes the other. Targeted items are then identified based on these item related features. A Kernel Density Estimation (KDE) method is used to calculate the burstiness in items. No review semantics is considered in identifying the connection between the reviewers.
Zhang \textit{et al.}~\cite{Zhang2020} propose a three step framework to address the limitations of the FIM model. In the first step, a graph of reviewers (nodes) is constructed based on the different relations (edges) such as correlation and relevance between the reviewers' ratings and their review times. In the next step, a label propagation algorithm based on propagation intensity and an automatic filtering mechanism is applied to obtain the final labels. A comparison of FIM-based and graph-based approaches is displayed in Fig.~\ref{fig:fim-vs-graph}.
\subsubsection{Research Gap} Previous approaches in both FIM and the graph-based methods extract handcrafted features which could be easily manipulated to deceive the detection approaches~\cite{Yuanshun2017}. Whereas, the word embedding from text written by a reviewer, has shown to contain more contextual embedding information than handcrafted features~\cite{Ren:2017:NND:3043977.3044107}. We recognize this in this research, and use word embeddings refined through a CNN as the reviewer representation. Also, the semantic relation between reviews from reviewers of a group is overlooked by the FIM and the graph-based approaches. In this research, we therefore devise a new graph representation learning neural network,
to 1) model temporal relationship (co-reviewing) between reviewers to predict the collaboration matrix, and 2) also to introduce the semantic collaboration through RNN. Finally, the collaboration networks are static and no network pruning or finetuning is considered. We recognize that a fraudster could camouflage in a genuine group and vice-versa, so pruning the least connected node helps mitigate this problem.

\section{Proposed Approach}
\label{sec:proposed-approach}
\begin{figure*}
\centering
\includegraphics[width=\linewidth]{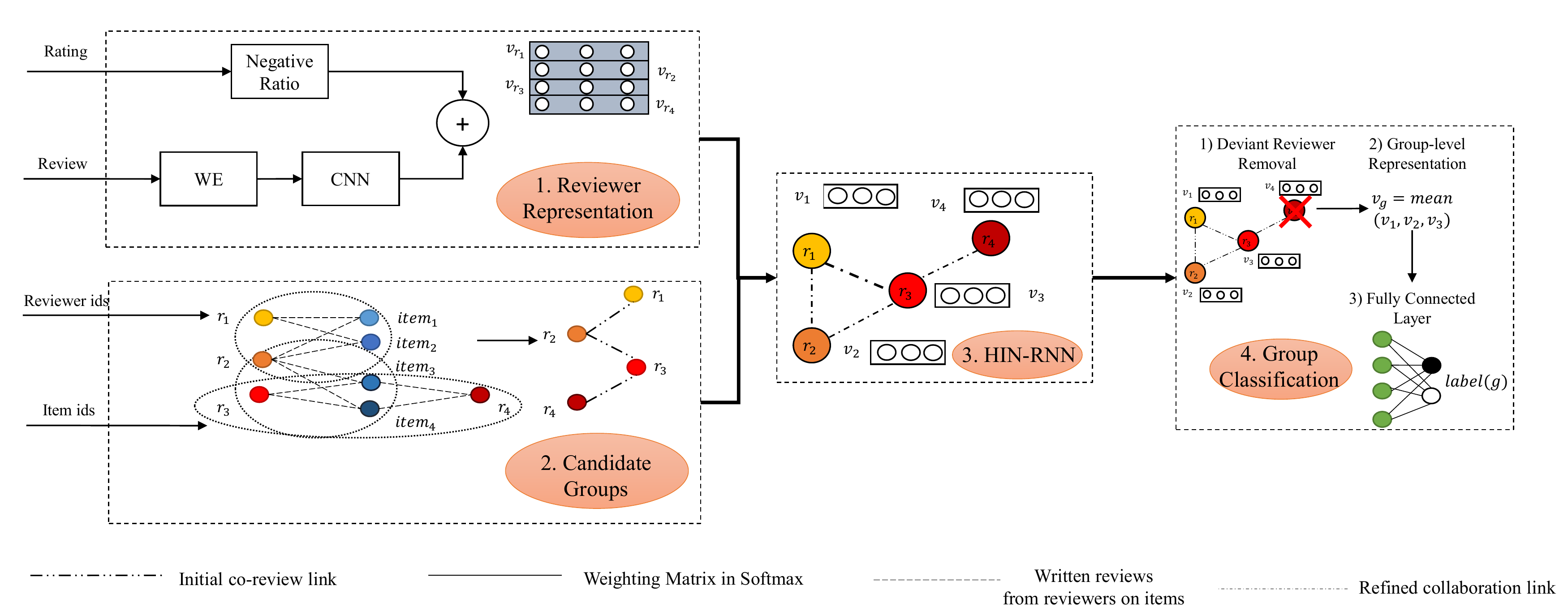}
\caption{The framework of our proposed approach. In this example the candidate group's initialization in step 2 (Candidate Groups), $r_1$ and $r_4$ have the minimum links. The HIN-RNN in step 3, refines the relationship according to the node types acquired from each reviewer semantic representation (step 1) and realizes a possible semantic collaboration between $r_1$ and $r_3$. On the other hand, the HIN-RNN recognizes no semantic relationship of $r_4$ with other reviewers. The $r_4$ remains the only reviewer with the minimum relationship. Hence, in step 4, $r_4$ is removed as a deviant reviewer.} 
\label{fig:framework}
\end{figure*}

\textit{\textbf{Problem Definition:}} Assume a set of review text ($T$) written by a a set of reviewers ($R$), for a set of items ($I$), with ratings ($rating$), the goal is to first, determine the groups ($G=\{g_1,...,g_i,...,g_m\}$ where $g\subset R$), and then classify the groups into $L_G=\{fraudster, genuine\}$.\\
In this research, we take advantage of the obviously and abundantly available, but often overlooked, review text as surrogates for reviewers, and establish relations between reviewers based on semantic similarities of their review text. Our proposed methodology therefore focuses on meaningful vector representation of reviewers and their relations obtained through four main modules as illustrated in Fig.~\ref{fig:framework}. 
A representation is obtained for each reviewer through aggregating written reviews in module 1, while initial candidate groups are created through co-review assumptions in module 2. 
The two vector representations are then combined using the HIN-RNN module to obtain a refined collaboration matrix. 
Finally, we exclude the deviant reviewers in a group to apply a simple fully connected layer to the average of the remaining reviewers' representation to output the final labels. 


\subsection{Reviewer Representation Extraction}
\label{sec:reviewer-representation}
Collective Intelligence (CI)~\cite{Malone2010,malone2009} states that the online footprints of a person is an important interpretation of his/her behavior. For example, a recruiter may look through a person's reviews online to understand his/her general behavior. Therefore a natural representation of a reviewer could be the collection of reviews. Traditional text-based features (e.g., the number of CAPITAL words in a review text, review text length) fail to represent a user and can easily be manipulated by fraudsters to fool the detectors. In contrast, Word Embedding (WE) techniques have shown a significant performance in fraud detection~\cite{Ren:2017:NND:3043977.3044107}. This is largely attributed to WE’s capabilities in capturing semantic similarities. In other words, review texts do not have to contain the same tokens to be deemed as similar by WE techniques. Sum of Word Embeddings (SoWEs) refers to an operation (in this case, addition) that aggregates word embeddings to produce a high-level description of a sentence, or a reviewer. Therefore, we use the SoWE at two different levels in this work: \textbf{reviewer-level} and \textbf{group-level}. Note, it is possible to use either sentence or document embedding techniques such as sec2vec, we choose SoWE due to its simplicity and proven efficiency in various domains~\cite{Ren:2017:NND:3043977.3044107,White2018,Lyndon2015}. The sentence-level embeddings are finetuned by a CNN with max-pooling. These aggregated representations are much more meaningful covering the global characteristics of each reviewer compared to handcrafted features as demonstrated in Sec.~\ref{sec:reviewer-rep-vs-group-features}. Extracting a \textbf{reviewer-level} representation follows three main steps of aggregations to cover the global characteristics of a reviewer. 

\subsubsection{Word Representation}
After aggregation of a reviewers' review texts, the reviews are broken down into sentences. Each sentence consists of words. In this step, each word in the sentence is initialized with a pre-trained word embedding represented as $e_{w_i}\in \mathcal{R}^D$, with $D$ as the word vector dimension. 
\subsubsection{Sentence Representation}
CNNs are capable of capturing the global characteristics of a sentence through three different layers~\cite{Zhang2016}. Here, a Convolutional Neural Network (CNN) is applied to refine the representations based on the reviewer types (genuine or fraudster). The CNN captures the reviewer type information through reviewer labels. So the CNN is fed with the word embedding and trained using the labels for each reviewer. The structure of the CNN we employed is given in Fig. \ref{fig:CNN}.
Since most reviews are less than 400 words, we selected the reviews with  less  than  400 words.  Then we pad words with “END”, so they can have a length of 400. In the first layer, a convolutional layer is utilized to operate as a tri-gram language model. The output of the first layer is given by:
\begin{equation}
\label{eq:conv-layer}
H_i = WI_{3,i} + b
\end{equation}
In Eq. \ref{eq:conv-layer}, $W\in \mathcal{R}^{1,3}$ is the weight matrix of the convolutional layer with stride $1$, $I_{3,i}\in\mathcal{R}^{3,D}$ is the concatenation of three consecutive words in the tri-gram model, and $b$ is the bias. Next, an average pooling is applied to the mean over the words of a sentence. 
Finally, a $tanh$ function is applied as an activation function. The activation function outputs the final embedding of a sentence: 
\begin{equation}
\label{eq:ave-pooling}
e_s = tanh(\frac{1}{n-2}\sum_{i=1}^{n-2}H_{i})
\end{equation}
In Eq.~\ref{eq:ave-pooling}, $n$ is the number of words in the sentence. 

\subsubsection{Reviewer Representation}
\begin{figure*}
\centering
\includegraphics[width=\linewidth]{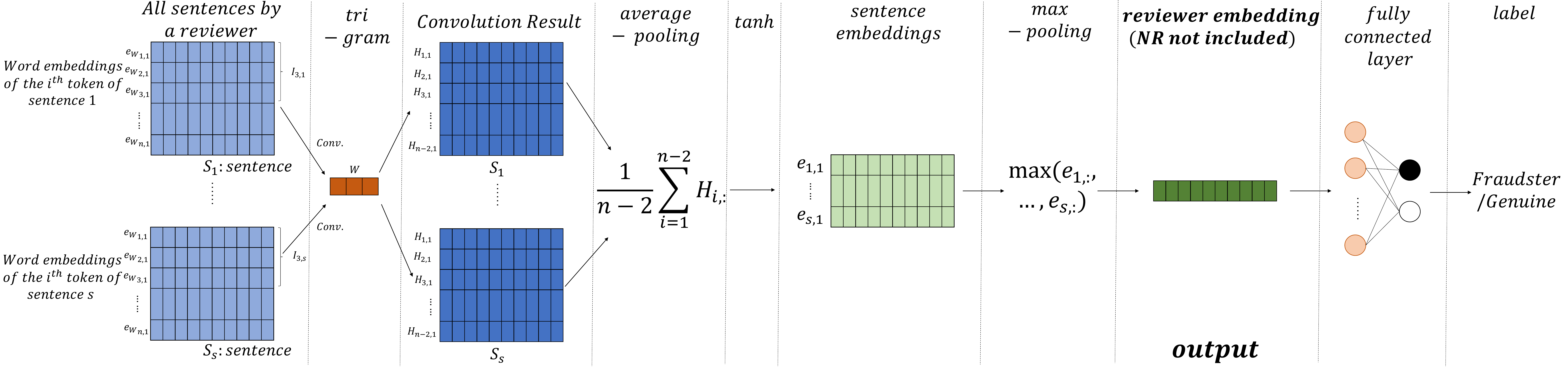}
\caption{The structure of our proposed CNN. In the first step of the CNN, the concatenation of word embeddings for a reviewer is fed to the network and the label of the reviewer is given in the final step. The output of the CNN is a reviewer embedding obtained before the fully connected layer.} 
\label{fig:CNN}
\end{figure*}

The reviewer representation is obtained by first concatenating the reviewer's sentence representations as a 2D tensor. Next, a max-pooling layer is applied to obtain the reviewer embedding, which is then fed to a fully connected layer to label the reviewer as a fraudster or genuine. The output of this step is the reviewer embedding for each reviewer (refer to Fig.~\ref{fig:CNN}). To train the CNN, we used a cross-entropy as the cost function.
Next, the embedding is concatenated with the Negative Ratio (NR)~\cite{Shebuit2015}:
\begin{equation}
\label{eq:nr}
NR = \frac{N(rating = 1, 2)}{N}
\end{equation}
where $N(rating)$ is the number of reviews with specific ratings ($rating$)  in the range  of  1-5  (5  is  the  highest),  and $N$ is  the  total number  of  reviews  for  each  reviewer. The concatenation forms the final vector representation of reviewer $r$ as $v_r$.

\subsection{Candidate Groups}
\label{sec:group-construction}
Fraudsters in a group tend to write their reviews in a shorter time-interval than genuine reviewers in an attempt to increase the collective impact. We assume a time-interval of 28 days similar to previous studies~\cite{Mukherjee2012,Mukherjee2013}. Based on this assumption, every two people who wrote reviews on a set of a minimum of two common items with the same rating in the specified time-window (in this case, 28 days) form a possible collaboration~\cite{JI2020454,Zhang2020,Wang2018}. The remaining reviewers are treated as individual reviewers and are ignored for fraudster group detection. The output of this step are subgraphs of possible groups, which we termed as the co-review networks.\\
\textit{\textbf{Definition:}} A co-review network is a subgraph $G(R,E)$ where $R$ is a set of reviewers, each reviewer ($r\in R$) as a node connected via an undirected edge ($e\in E$) with another reviewer if they have co-reviewed, and thus represent a possible collaboration.

\subsection{The HIN-RNN Framework}
\label{sec:representation-refinement}

As explained in the Introduction, fraudsters in a group develop a deep semantic relationship to cover the traces of the collaboration. To capture such relationships between two reviewers, the proposed model should be able to consider a long range relationship between the reviewers to improve the detection of candidate groups. Given the efficiency of an RNN in modeling long range dependencies we adopted an RNN to model the relationship between reviewers in a group. Our co-review networks contain two types of nodes, genuine reviewers and fraudsters, thus, strictly speaking, a Heterogeneous Information Network (HIN) or more precisely, in this case, a bipartite graph of two types of nodes. To adopt the representation learning in our co-review network, we propose HIN-RNN, an RNN compatible with HIN. A similar architecture, GraphRNN~\cite{You18a}, used an autoregressive model to generate graphs by training on a sequence of sub-graphs. 
Thus GraphRNN is capable of capturing the long term relationships between the nodes in a graph. However,  it does not support nodes of different types.

\subsubsection{GraphRNN} The adjacency matrix is a common way of representing a graph. Here we use a node ordering $\pi$ that maps the nodes to the rows and columns of an adjacency matrix. From here on we refer to this as the collaboration matrix  ($\pi(r_1), ..., \pi(r_i), ..., \pi(r_n)$). The $\Pi$ is a set of all possible permutations of the reviewers which is $n!$. If we pick a specific ordering $\pi\in\Pi$ the corresponding collaboration matrix is represented as $A^\pi\in \mathbb{R}^{n\times n}$ with $A_{i,j}^{\pi} = \mathbf{1}[\pi(r_i),\pi(r_j)\in E]$, where $E$ is the set of edges. The goal is to learn a set of distributions $p(G_i), G_i\in\{G_1, ...,G_i,...,G_s\}$ over all possible groups where each $G_i$ represents a grouping of reviewers within a collaboration matrix.\\
To map the problem to an autoregressive model, we need to define a mapping function: $f_S$ from co-review networks to a sequence. For a group $G\sim p(G)$ with $n$ different reviewers and $\pi$ as the ordering (possible collaboration between reviewers), the set of collaboration matrices for each reviewer is defined as below:
\begin{equation}
\label{eq:adj-seq}
S^{\pi} = f_S(G,\pi) = (S_1^\pi,S_2^\pi,...,S_n^\pi)
\end{equation}
As stated above, $S_i^\pi\in \{0,1\}^{i-1}, i\in\{1,...,n\}$ represents the collaboration vector ($S_i^\pi = (A_{1,i}^{\pi},...,A_{i-1,i}^{\pi})^T,\forall i \in \{2,...,n\}$) between reviewer $\pi(r_i)$ and previous reviewers $\pi(r_j),j\in \{1,...,i-1\}$ in a group. In other words, $S^\pi$ indicates a unique group representation $G$, where $f_G(S^\pi) = G$ ($f_G$ is the inverse of $f_S$). So, instead of learning $p(G)$, whose feature space is not easily characterized, the model learns $p(S^\pi)$, which is observable using the auxiliary function $\pi$. So, $p(G)$ can be represented as a joint distribution $p(G,S^\pi)$:
\begin{equation}
\label{eq:G-joint}
p(G) = \sum_{S^\pi}p(S^\pi)\mathbf{1}[f_G(S^\pi) = G]
\end{equation}
We are able to rewrite the $p(S^\pi)$ as an autoregressive conditional distribution:
\begin{equation}
\label{eq:autoregressive-col}
    p(S^\pi) = \prod_{i=1}^{n+1}p(S_i^\pi|S_{i}^\pi,S_{i-1}^\pi,...,S_{1}^\pi) = \prod_{i=1}^{n+1}p(S_i^\pi|S_{<i}^\pi)
\end{equation}
This means that the collaboration vector for each reviewer is dependent on the collaboration vectors on the previous reviewers. 

\subsubsection{HIN-RNN}
So far, the group generation process is modeled through the mapping of the possible collaborations into a collaboration vector for each reviewer. The collaboration vector for each reviewer is modeled through an autoregressive model which conditions each reviewer collaboration on all reviewers in a group. The GraphRNN learns a collaboration vector for each reviewer, but it is incapable of handling multiple reviewer types. In this research,
to include the reviewer types in generating the collaboration matrix, we condition the generation process on both the current reviewer's vector representation ($v_r$ from Sec. \ref{sec:reviewer-representation}) and the collaboration vectors of previous reviewers:
\begin{equation}
\label{eq:autoregressive-edge}
p(S_i^\pi|S_{<i}^\pi) = \prod_{j=1}^{i-1}p(S_{i,j}^\pi|v_i, S_{i,<j}^\pi,S_{<i}^\pi)
\end{equation}
\begin{figure*}
\centering
\includegraphics[width=\linewidth]{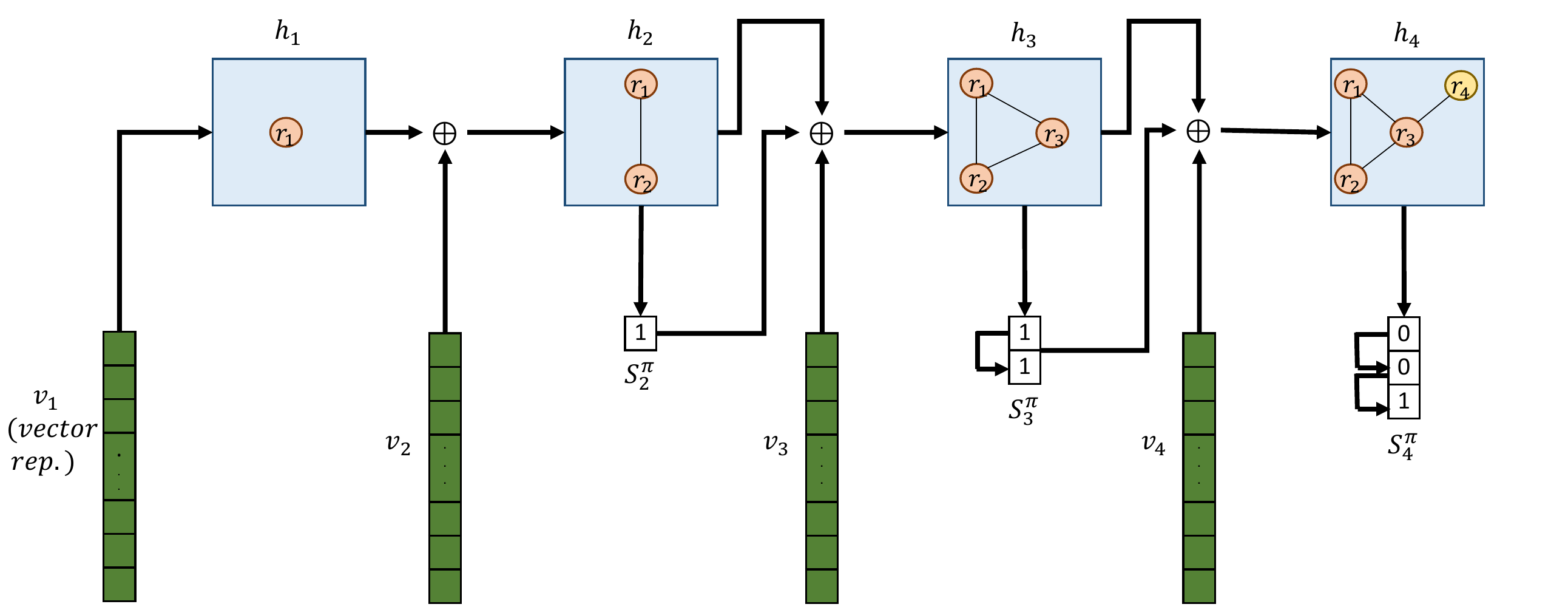}
\caption{The structure of the HIN-RNN. The HIN-RNN uses the representation of each reviewer, the graph state, and the adjacency matrix of the previous node to generate the adjacency matrix for the current reviewer. Note that $\oplus$ is the concatenation operator.} 
\label{fig:HIN-RNN}
\end{figure*}

In Eq. \ref{eq:autoregressive-edge} $S_{i,j}^\pi$ is 1 if there is a collaboration between reviewer $i,j$. 
The structure of the HIN-RNN model is displayed in Fig. \ref{fig:HIN-RNN}.
Two models are required to the first to parameterize how reviewers are connected in the collaboration matrix, and the second to how previous reviewers are interconnected to each other. For each model we use a Recurrent Neural Network (RNN) to capture the distributions:
\begin{equation}
h_i = f_{1}(h_{i-1},S_{i-1}^\pi,v_{i})
\end{equation}
where $h_i$ encodes the state of the groups (reviewers plus their collaboration matrix) up to reviewer $i$ and $f_1$ is the function learnt through training the RNN. Next, we obtain the collaboration matrix of the current reviewer: 
\begin{equation}
S_i^\pi = f_{2}(h_i)
\end{equation}
where $S_i^\pi$ encodes the collaboration matrix obtained from the function $f_2$ using the RNN. 

\subsection{Group Classification}
\label{sec:classification}
{Our proposed group classification method follows three simple steps: deviant reviewer removal, group-level representation, and fully-connected layer.} 
\subsubsection{Deviant Reviewer Removal}
In this step, from the groups generated at the output of the HIN-RNN described in
Sec. \ref{sec:representation-refinement}, 
we exclude the reviewers with a minimum collaboration in a group to remove the possibility of unintentional contribution from genuine reviewers in a fraud activity (reducing FP) and avoid the effects of fraudulent activity in a genuine group (reducing FN). The final output of this step includes groups with deviant reviewers excluded. 
\subsubsection{Group-Level Representation}
\label{sec:group-level}
Next, we extract a \textbf{group-level} representation (refer to Sec. \ref{sec:reviewer-representation}) using the CI theory. Finally we perform an element-wise average over the representations of the remaining reviewers to obtain a final representation of each group. The element-wise average of a group's representation is the output of this step.
\subsubsection{Fully-Connected Layer}
\label{sec:fc-layer}
In this step, a fully connected layer is trained on the representation obtained from Sec. \ref{sec:group-level}. \\
The proposed approach is presented in Algorithm \ref{alg:overall-alg}.

\begin{algorithm}[hbt!]
 \caption{Proposed Algorithm}
\label{alg:overall-alg}
\textbf{Output:}\
The label of each group\;
\textbf{Input:}\
$T$ review texts of $R$ reviewers and ratings $rate$, written on $I$ items\;

 \% \textbf{Step 1:} reviewer representation\;
 {
    \For{$r \gets 1$ to $R$} 
    {
        \% Aggregate reviews of reviewer $r$\;
        $t_r \leftarrow aggregate\{T_1, T_2, ..., T_m\}$\;
        \% Tokenize $t_r$ to $S$ sentences\;
        $\{s_1,s_2,...s_S\} \leftarrow tokenize(t_r)$\; 
        \% Sentence Representation\;
        \For{$s \gets 1$ to $S$} 
        {
            \% Tokenize $S_s$ to $n$ words\;
            $\{w_1, w_2, ..., w_n\} \leftarrow tokenize(S_s)$\; 
            \% Word embeddings\;
            $\{e_{w_1}, e_{w_2}, ..., e_{w_n}\} \leftarrow WE(\{w_1, w_2, ..., w_n\})$\;
            \% The sentence representation\;
            $e_{s} \leftarrow CNN(\{e_{w_1}, e_{w_2}, ..., e_{w_n}\})$\; 
        }
        \% Negative Ratio of reviewer $r$\;
        $NR_r\leftarrow NR(rate_1,rate_2,...,rate_m)$\;
        \% Final representation of $r$\;
        $v_r \leftarrow maxPool(concat({e_{s},\forall s\in S})) \oplus NR_r$\; 
    }
}
 \% \textbf{Step 2:} candidate groups subgraphs\;
            \uIf{$r_x,r_y$ co-review same item $i\in I, \forall r_x,r_y\in R$}{
            \% Link the possible collaborating reviewers\;
                $E(x,y)$ = 1; 
            }
\% \textbf{Step 3:} final collaboration matrix using HIN-RNN\; 
    \For{$i \gets 1$ to $R$}
    {
        $h_{r_i} \leftarrow RNN_1(h_{r_{i-1}},E(0:i-1,0:i-1),v_i)$\;
        \% Collaboration matrix of reviewer $i$\;
        $C_i \leftarrow  RNN_2(h_{r_i})$; 
    }
    $G \leftarrow $ reviewers with same connected reviewers in $C$\;

\% \textbf{Step 4:} group classification\; 
    \For{$g \in G$}
    {
        $g \leftarrow g - $\{reviewer with minimum connections\} \;
        $v_g\leftarrow mean(v_r)$ for $r\in g$\;
        $label(g)\leftarrow fc(v_g)$\;
    }

\end{algorithm}


\section{Experimental Evaluation}
\label{sec:experimental-evaluation}
We compare the proposed approach to the state-of-the-art approaches to demonstrate the effectiveness of each of our innovations.  
\subsection{Experimental Setup}
\label{sec:epxerimental-setup}
We used a 100-dimension Continuous Bag of Words (CBoW) due to its effectiveness in fraud review detection \cite{Zhang2016} with a window size of 2 and batch size of 256. For training the CNN, the learning rate was $10^{-4}$, the training epochs was 30, and cross-entropy was used as the objective function. To train the group generator model, we used two RNNs with Gated Recurrent Units (GRUs), one learns the hidden state of the group with a hidden size of 128, and the other learns the reviewer collaboration with a hidden size of 16. The two RNNs were trained jointly with a learning rate of 0.003 and in 3000 epochs.   

\subsection{Datasets}
\label{sec:datasets}
\begin{center}
\begin{table}[hbt!]
\centering
\caption{List of datasets used in our current study.}\label{tab:datasets}
\begin{tabular}{|c|c|c|c|c|}
\hline
Dataset &  Reviewers & Items & Reviews & Candidate Groups\\
\hline
\hline
Yelp &  260,277 & 5,044 & 608,598 & 9,952\\
Amazon &  42,655 & 6,822 & 53,777 & 2,194\\
\hline
\end{tabular}
\end{table}
\begin{center}
\begin{table*}
\centering
\caption{A comparison between the results on the proposed approach and two state-of-the-art systems \cite{Zhang2020,JI2020454}.}\label{tab:comparison}
\begin{tabular}{|c|c|c|c|c|c|c|}
\hline
Dataset & \multicolumn{3}{c|}{Yelp} & \multicolumn{3}{c|}{Amazon} \\
\hline
\hline
Metric &  Precision & Recall & F1-value & Precision & Recall & F1-value\\
\hline
Zhang \textit{et al.} \cite{Zhang2020}&  0.70 & 0.20 & 0.32 & 0.80 & 0.45 & 0.58 \\
Ji \textit{et al.} \cite{JI2020454} &  \textbf{0.83} & 0.60 & 0.69 & 0.82 & \textbf{0.92} & 0.86 \\
{GraphRNN + SoWE + NR} & {0.77} & {0.72} & {0.75} & {0.81} & {0.88} & {0.84} \\
HIN-RNN + SoWE + NR & 0.81 & \textbf{0.82} & \textbf{0.81} & \textbf{0.85} & 0.90 & \textbf{0.87} \\
\hline
\end{tabular}
\end{table*}
\end{center}

\end{center}
Previous studies evaluated the performance of the proposed approach on either the Yelp or Amazon dataset. To demonstrate the scalability of the proposed approach, we used both datasets. The Yelp dataset includes reviews from ``20-Oct-2004" to ``10-Jan-2015" and provides the labels (genuine or fraudster) for possible groups~\cite{Zhang2020}. The Amazon dataset contains reviews from ``01-Feb-2000" to ``10-Oct-2010", and groups are determined and labeled, similar to the Yelp dataset~\cite{JI2020454,Zhang2020}. In both datasets, the fraudsters who have co-reviewed the same set of items with others might form fraudster groups. 
The details on the datasets are provided in Table \ref{tab:datasets}. For the evaluation, we used 80\% of the data for the training set and 20\% of the data for the test set.

\subsection{Main Results}
\label{sec:main-results}
The performance of our proposed approach is measured using three metrics: precision ($\frac{TP}{TP + FP}$), recall ($\frac{TP}{TP + FN}$),  and F1-value ($\frac{2\times precision\times recall}{precision + recall}$).\\
\subsubsection{Comparison with baselines}
\label{sec:comparison}
The performance of the proposed approach is compared against two state-of-the-art systems in Table  \ref{tab:comparison}. 
We also devised two different network configurations to compare the performance of the HIN-RNN with other graph generation techniques:\\
\textbf{GraphRNN + SoWE + NR:} In this configuration, the reviewer representation (SoWE + NR) is not employed to refine the groups' structure. Though, the reviewer representation is used to obtain the group-level representation. \\
\textbf{HIN-RNN + (SoWE + NR):} This is the regular proposed configuration where features are employed to refine the group structure.\\
The results demonstrate that the proposed approach significantly outperforms the baselines in terms of recall and F1-value on the Yelp dataset. The proposed approach also improves the baselines' performance in terms of Precision and F1-value. 
As mentioned in Sec. \ref{sec:related-works}, Ji \textit{et al.} \cite{JI2020454} provide a strategy to overcome the limitations of previous works in excluding genuine reviewers who unintentionally assigned to a fraudster group. However, it still suffered from a limitation in handling such situations. So FN is not reduced significantly (i.e. recall is low) compared to Zhang \textit{et al.} \cite{Zhang2020}. On the other hand, the approach by Ji \textit{et al.} performs better when faced with single fraudsters who co-review the non-targeted items with groups of genuine reviewers, and thus decreases FP (and increases precision). 
The GraphRNN, although improving the performance, still suffers from overlooking the semantic relationship between reviewers in a group.
The HIN-RNN effectively increases the model flexibility to identify deviant reviewers' involvement in a group and then remove them. 
Hence, both FP and FN are significantly reduced, resulting in a \textbf{22\%} improvement in recall and \textbf{12\%} in F1-value on the Yelp dataset. \\
\subsubsection{Effects of the reviewers' representation vs. handcrafted individual features}
\label{sec:reviewer-rep-vs-group-features}
\begin{figure}[h]
\centering
\includegraphics[width=0.47\textwidth]{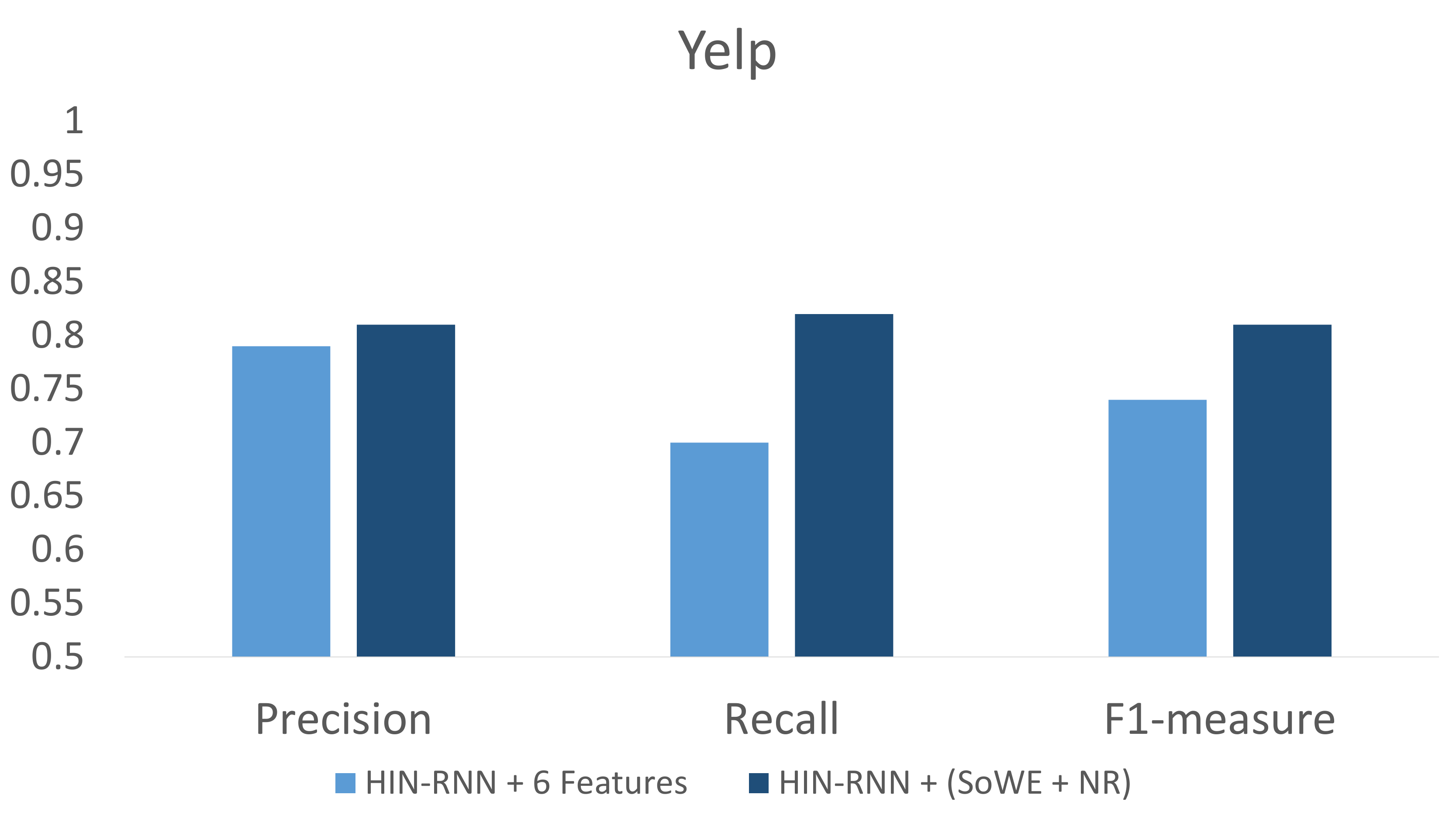}
\caption{The performance of SoWE + NR against handcrafted features on the Yelp dataset.} 
\label{fig:features-yelp}
\end{figure}
\begin{figure}[h]
\centering
\includegraphics[width=0.47\textwidth]{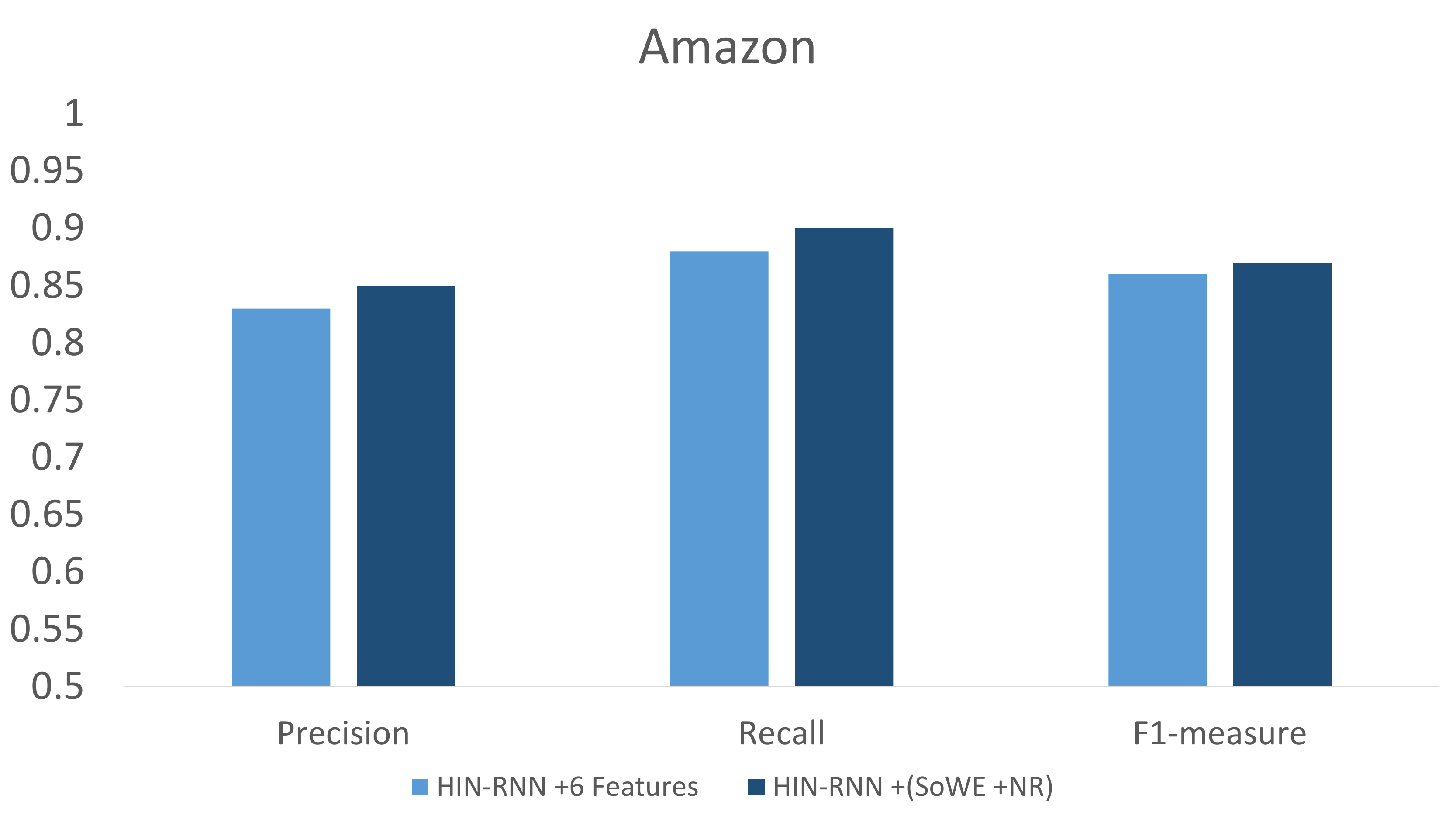}
\caption{The performance of SoWE + NR against handcrafted features on the Amazon dataset.} 
\label{fig:features-amazon}
\end{figure}
\begin{figure*}
\begin{subfigure}{0.46\textwidth}
\centering
\includegraphics[width=\textwidth]{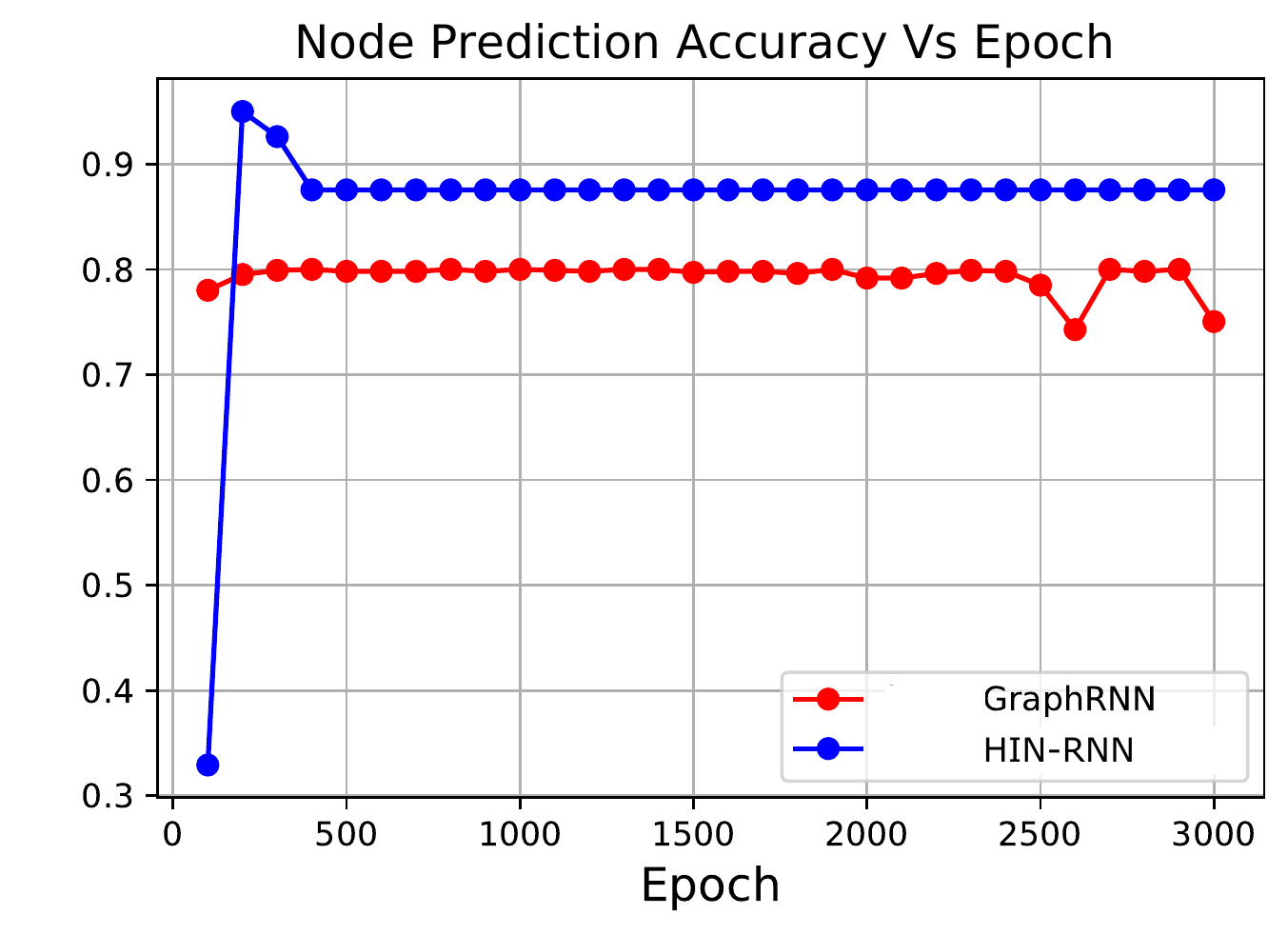}
\caption{Node prediction accuracy vs training epochs on Yelp.} 
\label{fig:node_yelp}
\end{subfigure}
\hfill
\begin{subfigure}{0.46\textwidth}
\centering
\includegraphics[width=\textwidth]{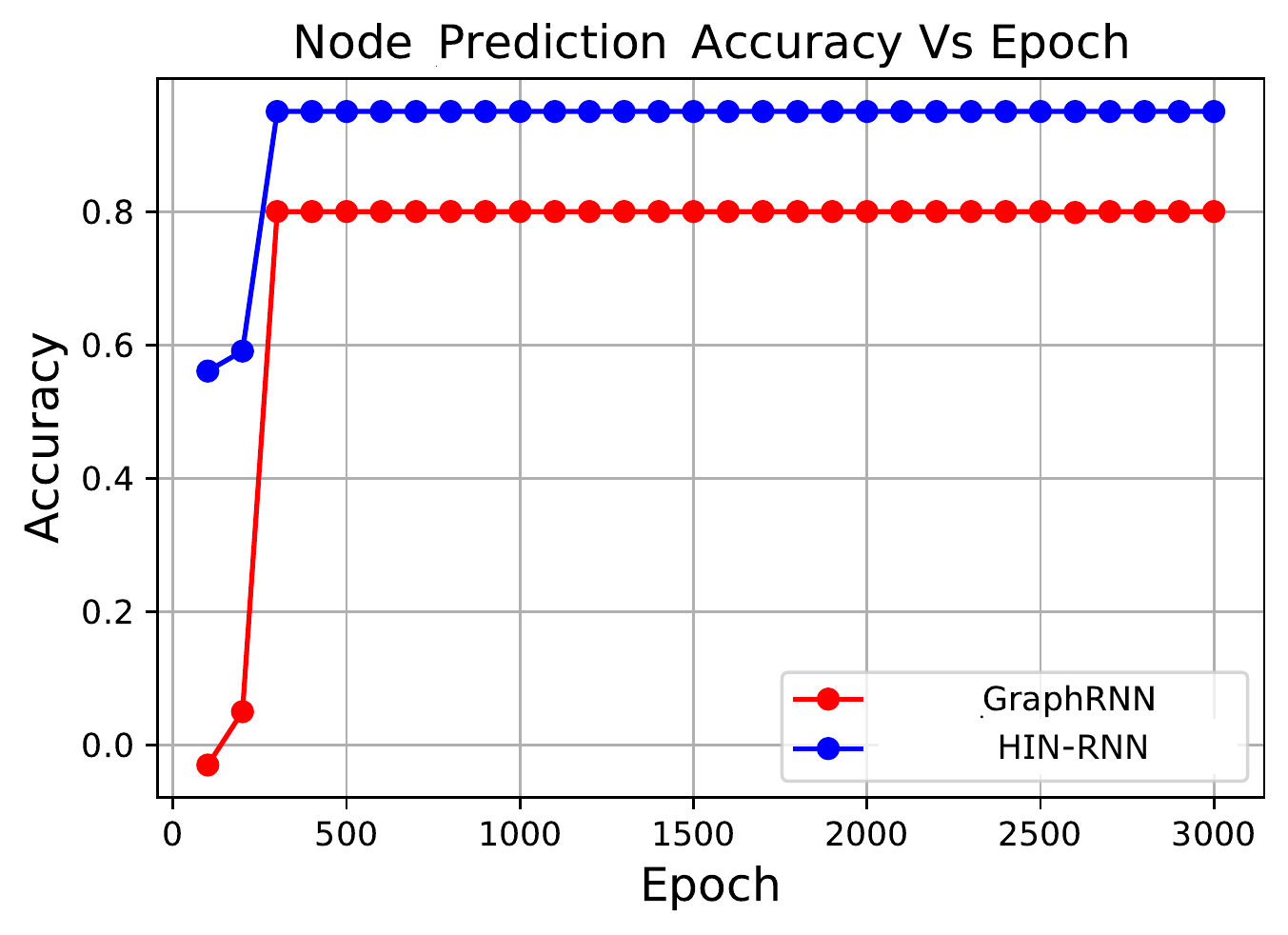}
\caption{Node prediction accuracy vs training epochs on Amazon.} 
\label{fig:node_amazon}
\end{subfigure}
\hfill
\begin{subfigure}{0.46\textwidth}
\centering
\includegraphics[width=\textwidth]{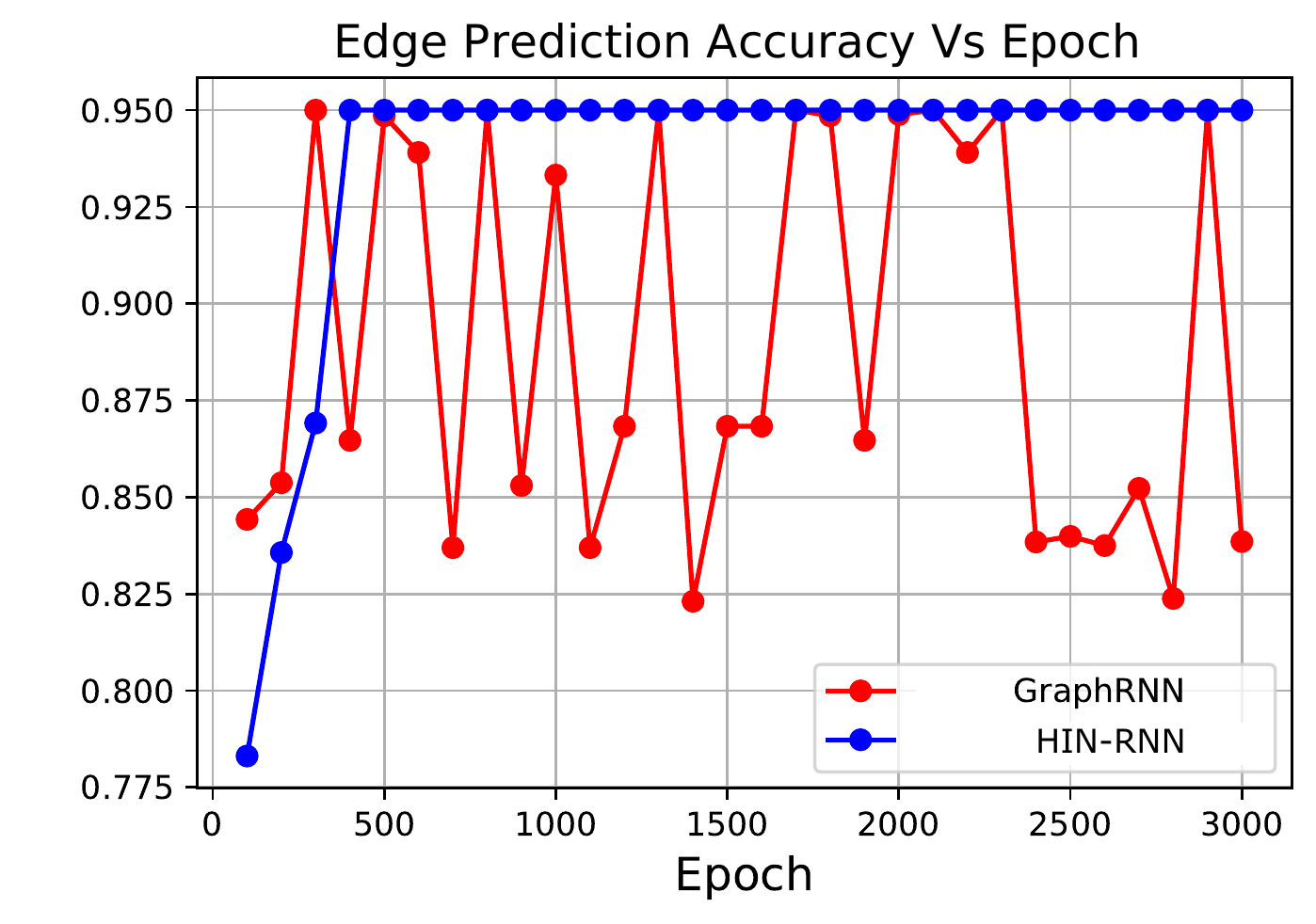}
\caption{Edge prediction accuracy vs training epochs on Yelp.} 
\label{fig:edge_yelp}
\end{subfigure}
\hfill
\begin{subfigure}{0.46\textwidth}
\centering
\includegraphics[width=\textwidth]{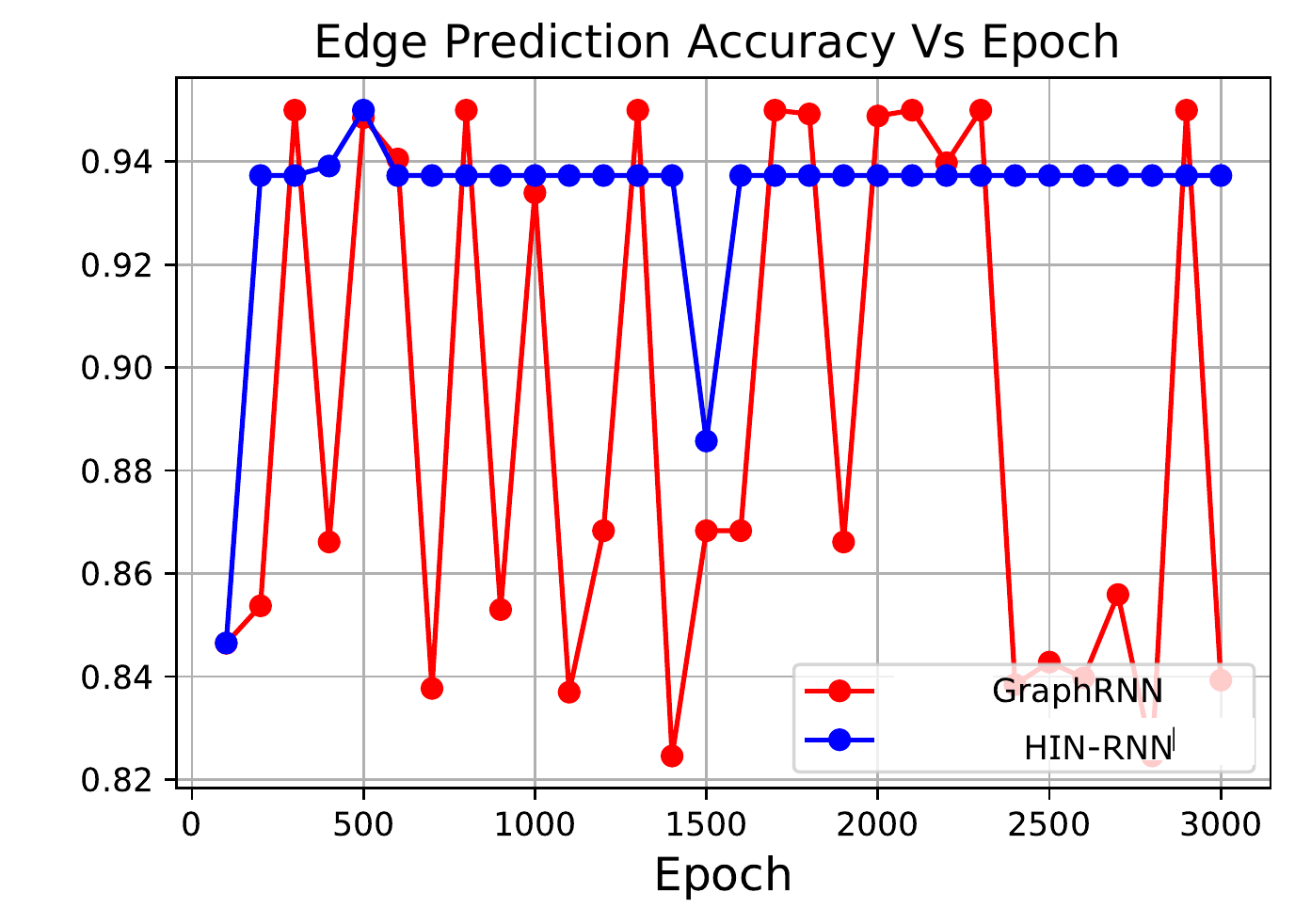}
\caption{Edge prediction accuracy vs training epochs on Amazon.} 
\label{fig:edge_amazon}
\end{subfigure}
\caption{HIN-RNN performance on edge and node prediction vs number of training epochs.}
\label{fig:node_edge_accuracy}
\end{figure*}
Ji \textit{et al.} \cite{JI2020454} incorporate 6 handcrafted individual features (discussed in Sec. \ref{sec:related-works}). Two versions of our proposed approach are presented here to show the effectiveness of the proposed features:\\ 
\textbf{HIN-RNN + 6 features from Ji \textit{et al.}~\cite{JI2020454}:} This configuration utilizes the 6 individual fraudster features proposed by Ji \textit{et al.} as discussed in Sec.~\ref{sec:related-works}. In this configuration each reviewer is represented as a concatenation of the following handcrafted features: Ratio of Extreme Rating, Rating Deviation, The most Reviews One-day, Review Time Interval, Account Duration, and Active Time Interval Reviews. We then use the handcrafted features to refine the candidate groups.\\ 
\textbf{HIN-RNN + (SoWE + NR):} This configuration utilizes SoWE as the feature representation (See Sec. \ref{sec:reviewer-representation}.)
As shown in Figs. \ref{fig:features-yelp} and \ref{fig:features-amazon} the handcrafted features can significantly affect the performance, since these features fail to capture a comprehensive representation of groups. Such features mostly rely on simple statistics of a group such as Group Size or Group Rating Deviation. The semantics of the written reviews by reviewers are overlooked, despite its proven effectiveness~\cite{Shebuit2015,Zhang2016}. We not only use the semantics in the \textbf{reviewer-level} representation but also a \textbf{group-level} representation of each group after excluding the deviant reviewers from the groups. 
Employing SoWE to represent reviewers results in capturing a global representation of reviewers, and consequently improves the fraudster group detection. 
\subsubsection{HIN-RNN performance in graph generation}
\label{sec:representation-effect}

\begin{figure*}
\begin{subfigure}{0.46\textwidth}
\centering
\includegraphics[width=\textwidth]{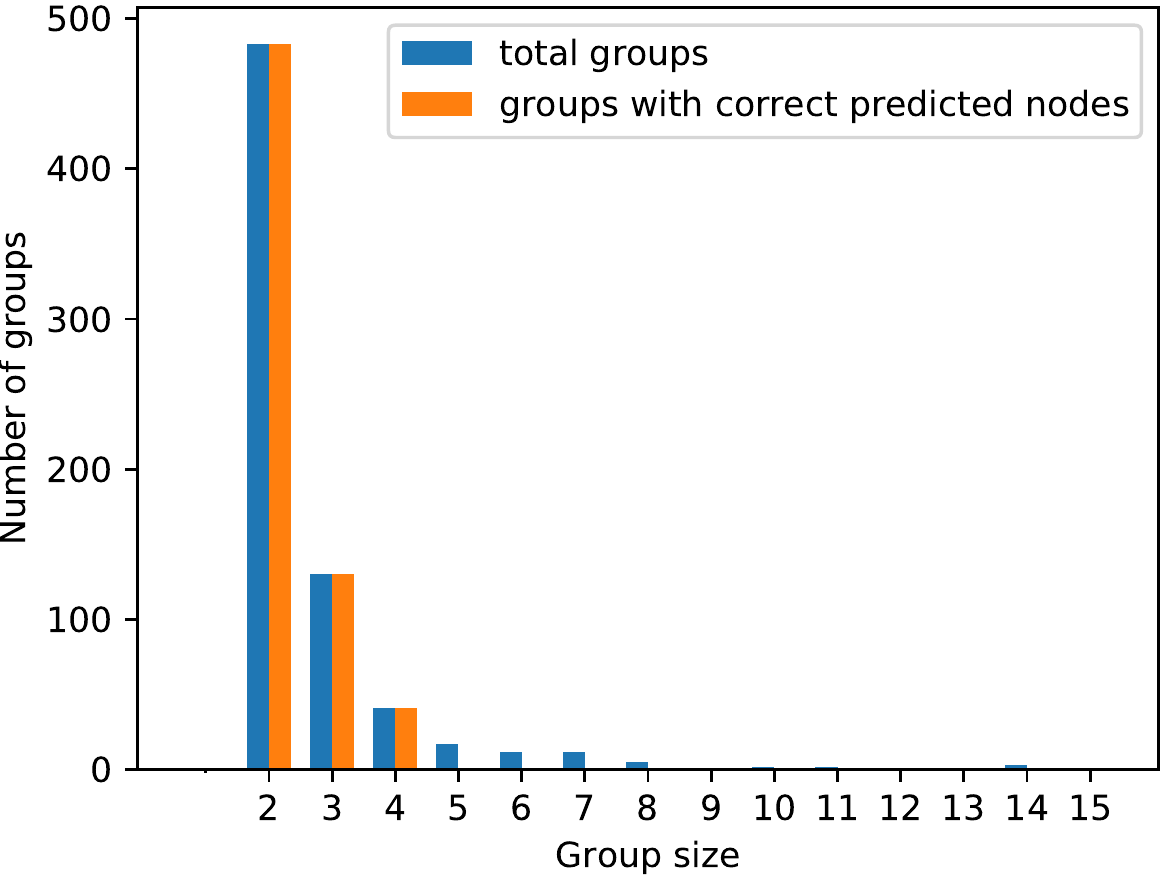}
\caption{Node prediction accuracy vs group size on Yelp.} 
\label{fig:node_yelp_size}
\end{subfigure}
\hfill
\begin{subfigure}{0.46\textwidth}
\centering
\includegraphics[width=\textwidth]{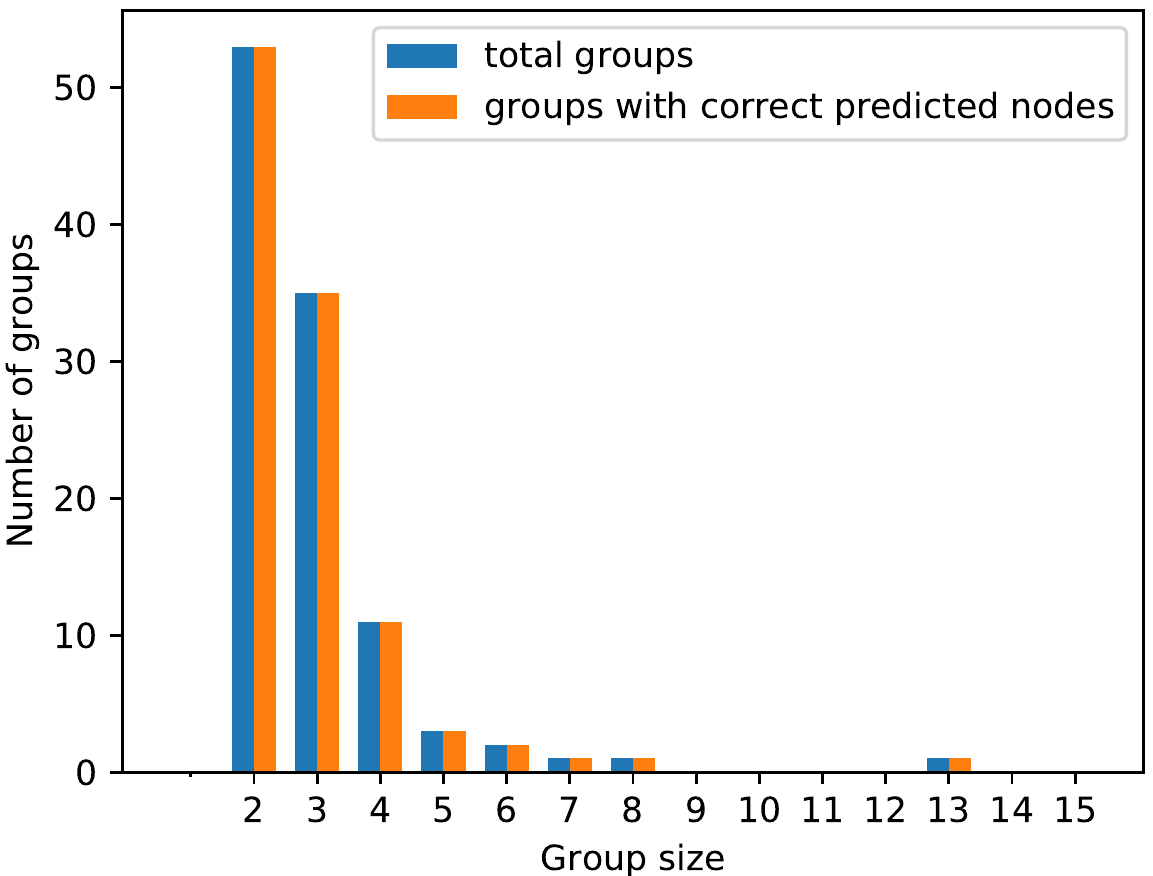}
\caption{Node prediction accuracy vs group size on Amazon.} 
\label{fig:node_amazon_size}
\end{subfigure}
\hfill
\begin{subfigure}{0.46\textwidth}
\centering
\includegraphics[width=\textwidth]{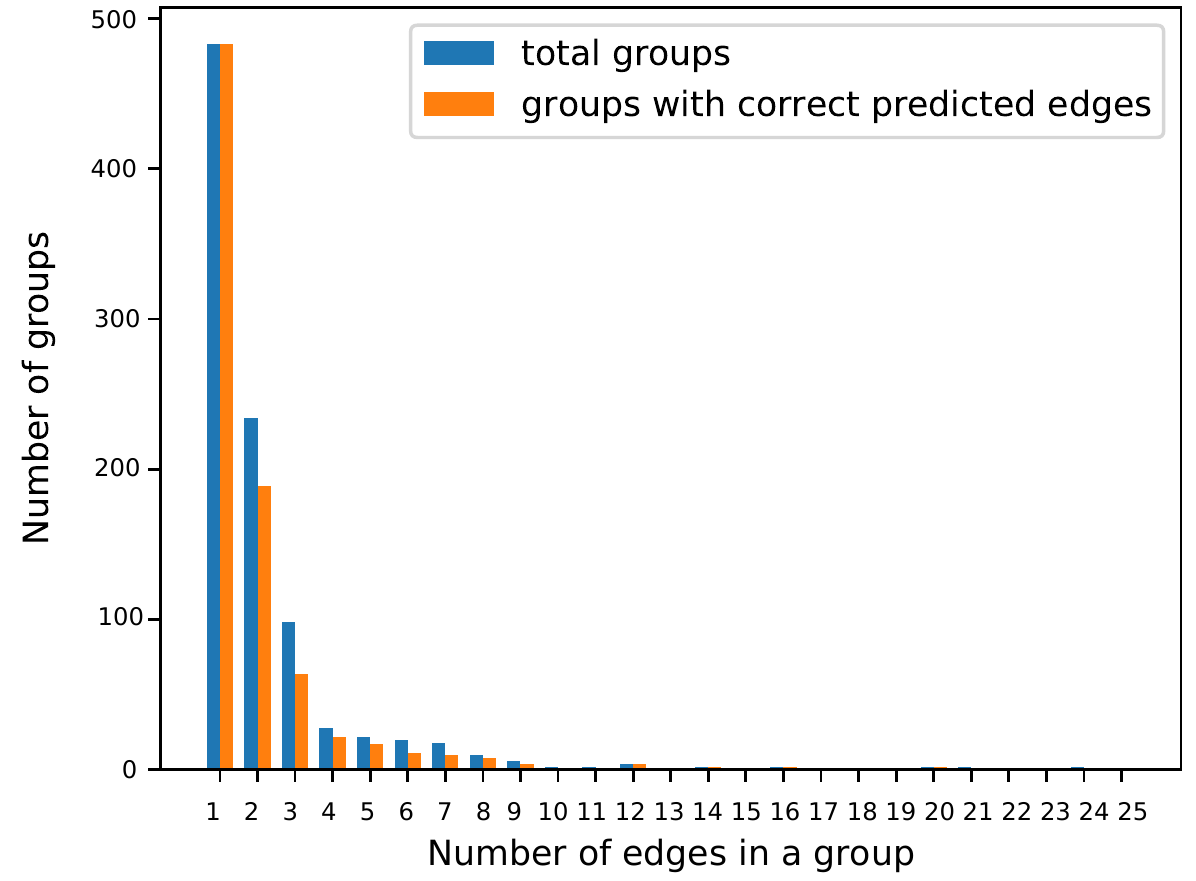}
\caption{Edge prediction accuracy vs group size on Yelp.} 
\label{fig:edge_yelp_size}
\end{subfigure}
\hfill
\begin{subfigure}{0.46\textwidth}
\centering
\includegraphics[width=\textwidth]{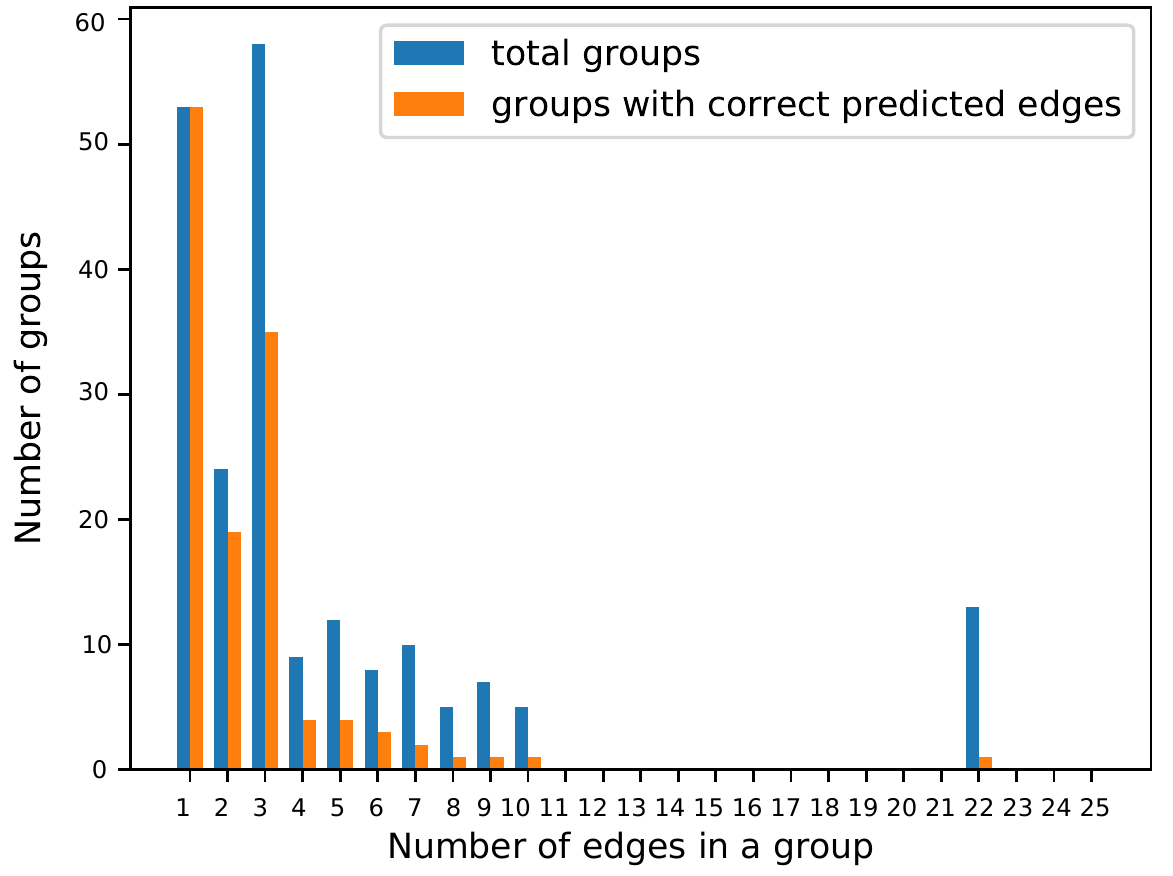}
\caption{Edge prediction accuracy vs group size on Amazon.} 
\label{fig:edge_amazon_size}
\end{subfigure}
\caption{HIN-RNN performance on edge and node prediction vs group size and number of edges in each group.}
\label{fig:node_edge_accuracy_size}
\end{figure*}

To evaluate the performance of the HIN-RNN, we defined accuracy as a ratio of the number of correctly predicted edges $|E_c|$ over the total number of edges $E$:
\begin{equation}
    \label{eq:node-edge-acc}
    Acc. = \frac{|E_c|}{|E|}
\end{equation}
We use accuracy to measure the performance of the HIN-RNN for the graph generation task. 
Figs. \ref{fig:node_yelp}, \ref{fig:node_amazon}, \ref{fig:edge_yelp}, and \ref{fig:edge_amazon} show the accuracy on the node and edge prediction compared to the number of training epochs on the test set for both the Yelp and Amazon datasets. The plots demonstrate the effectiveness of the HIN-RNN (blue) in predicting the nodes and the edges compared to the GraphRNN (red). The results show that including the node type in the prediction through each nodes' (reviewer representation in current study) representation improves the node prediction by an average of 8\% (Fig. \ref{fig:node_yelp}) on the Yelp dataset and 12\% (Fig. \ref{fig:node_amazon}) on the Amazon dataset. In other words, including the representation of a reviewer considers the overlooked deep semantic contribution to the collaboration between the reviewers and hence improves the model to better predict reviewers in a group. 

The edge prediction accuracy is also improved over the GraphRNN with an average of 15\% on the Yelp dataset and an average of 8\% on the Amazon dataset. Training the edges with nodes' representation enables the HIN-RNN to rely on a corresponding feature space (acquired from the reviewer representation) to predict the possible collaboration. So two reviewers with similar representations and belonging to the same type have the same collaboration matrix, while two reviewers with different representations (and hence belong to different types) are unlikely to collaborate, even if two reviewers share the same connections as the candidate groups. 
The same trend also explains the fluctuation in training epochs.

We also conducted experimental evaluations on the number of correct nodes and the number of correct edges in groups with different sizes to show the effectiveness of the HIN-RNN in graph generation with respect to the graph complexity (size, number of edges) of the group. Fig.~\ref{fig:node_edge_accuracy_size} shows the performance of the HIN-RNN on edge and node prediction against the group size and the number of edges for the Yelp and Amazon datasets. As Fig. \ref{fig:node_yelp_size} shows the HIN-RNN correctly predicts the nodes for smaller groups (up to 4 nodes) on the Yelp dataset, but as the number of nodes in a group increases the number of correctly predicted nodes decreases. On the other hand, the HIN-RNN correctly predicts all the nodes for different sizes on the Amazon dataset as shown in Fig. \ref{fig:node_amazon_size}. We also displayed the performance of edge prediction in Fig. \ref{fig:node_edge_accuracy_size}. Fig. \ref{fig:edge_yelp_size} shows the performance of the HIN-RNN on the Yelp dataset for groups of different edge numbers. The HIN-RNN performance stays stable across groups with a different number of edges. Hence, we can claim that the HIN-RNN is capable of capturing the long range dependencies between the nodes in a group. As the number of edges increases, the HIN-RNN can still reliably model the relationships. Although the number of predicted edges on groups with a higher number of edges decreases with the Amazon dataset, the HIN-RNN is still capable of capturing the relationships between the nodes in a group with a higher number of edges, e.g., 22 edges.
\subsubsection{Effects of reviewers removal}
\label{sec:reviewer-removal-effect}

\begin{figure}[h]
\centering
\includegraphics[width=0.3\textwidth]{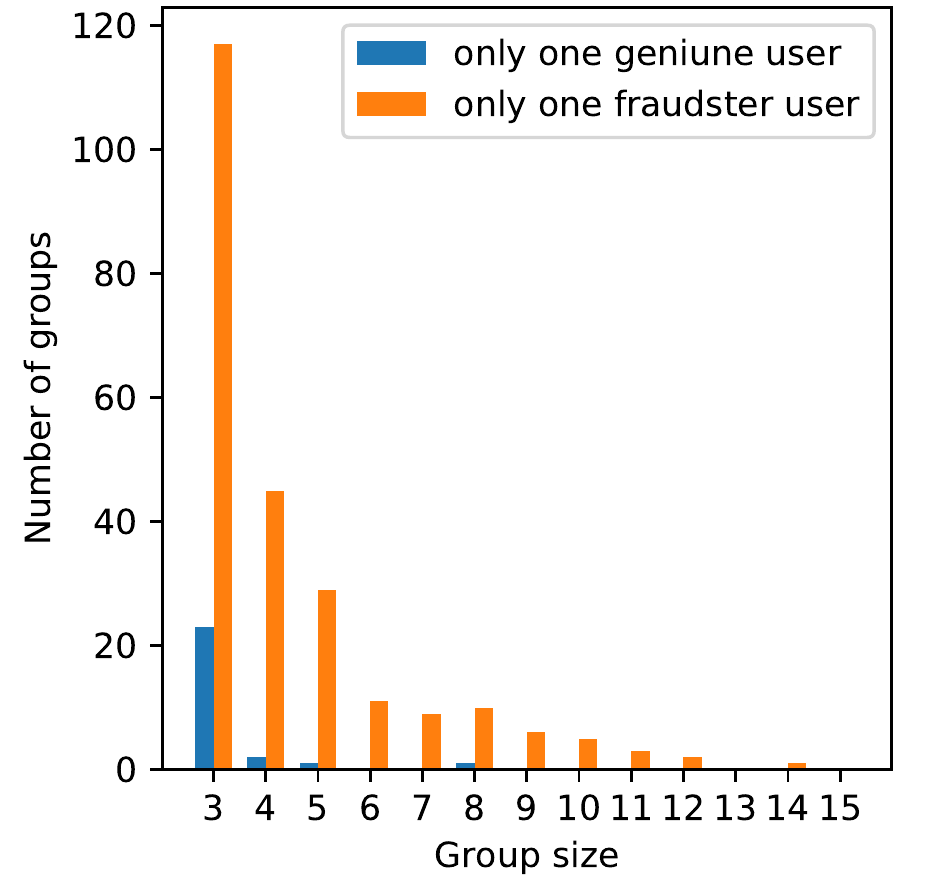}
\caption{Distribution of groups with only one fraudster or only one genuine reviewer vs. the group size on the Yelp dataset.} 
\label{fig:imposter-yelp}
\end{figure}
\begin{figure}[h]
\centering
\includegraphics[width=0.3\textwidth]{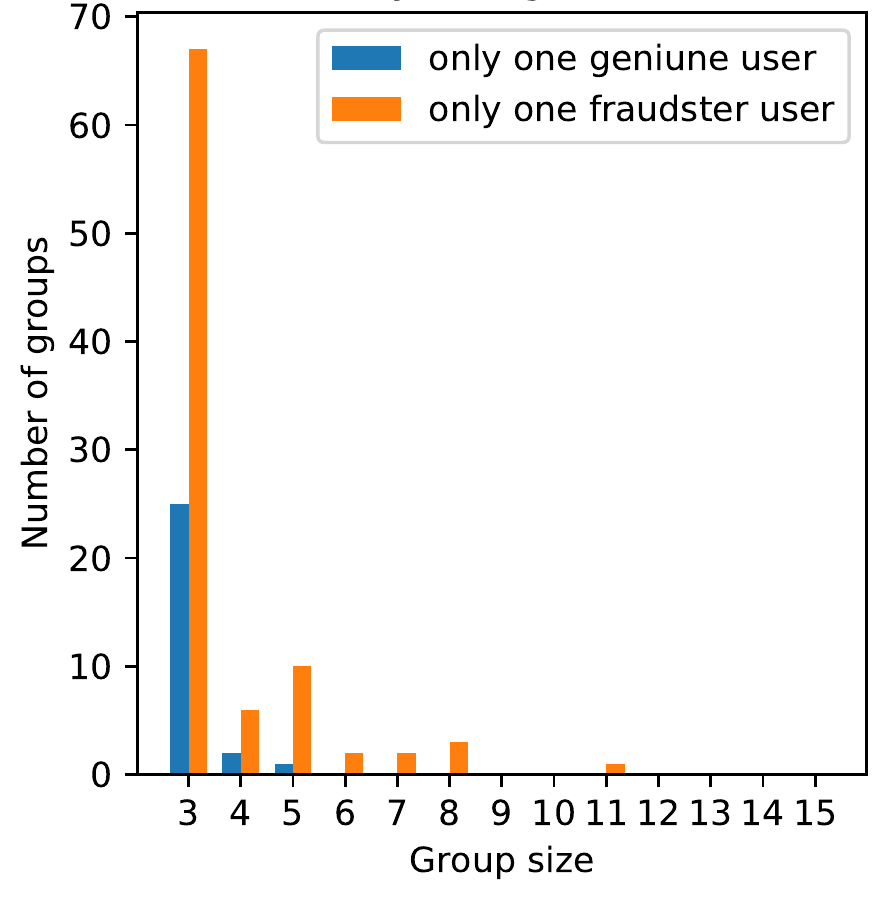}
\caption{Distribution of groups with only one fraudster or only one genuine reviewer vs. the group size on the Amazon dataset.} 
\label{fig:imposter-amazon}
\end{figure}

As explained, genuine reviewers are in some cases involved in unwanted fraudster group activities, and on the other hand, fraudsters may also camouflage themselves in genuine groups to escape detection. Figs. \ref{fig:imposter-yelp} and \ref{fig:imposter-amazon} show the distribution of the groups with only one fraudster or one genuine reviewer against the group size. As Fig. \ref{fig:imposter-yelp} shows, the number of groups with only the one fraudster is much higher than groups with only one genuine reviewer. Additionally, single fraudsters are camouflaged in different groups regardless of the group size. This is intuitively possible since single fraudsters employ camouflage in any possible situation to escape detection. Genuine reviewers, on the other hand, contribute less to fraudster group activity and they mainly co-review with two fraudsters in a group with a size of 3 reviewers. But as Fig. \ref{fig:imposter-yelp} shows, genuine reviewers' involvement is significantly decreased, as the group size grows. The Amazon dataset shows a similar trend in Fig. \ref{fig:imposter-amazon}. Genuine reviewers are similarly involved in a fraudster activity for a group with sizes of 3, 4, and 5, but fraudsters are involved in genuine groups across a wider range of group sizes. The single genuine reviewer in a fraudster group decreases the fraudster score of the group, which in turn increases False Negative (FN) labeled samples. On the other hand, with fraudsters camouflaged in a genuine group, the fraudster score of a genuine group increases. This leads to an increase in False Positive (FP) labeled samples. 
\begin{figure}[h]
\centering
\includegraphics[width=0.47\textwidth]{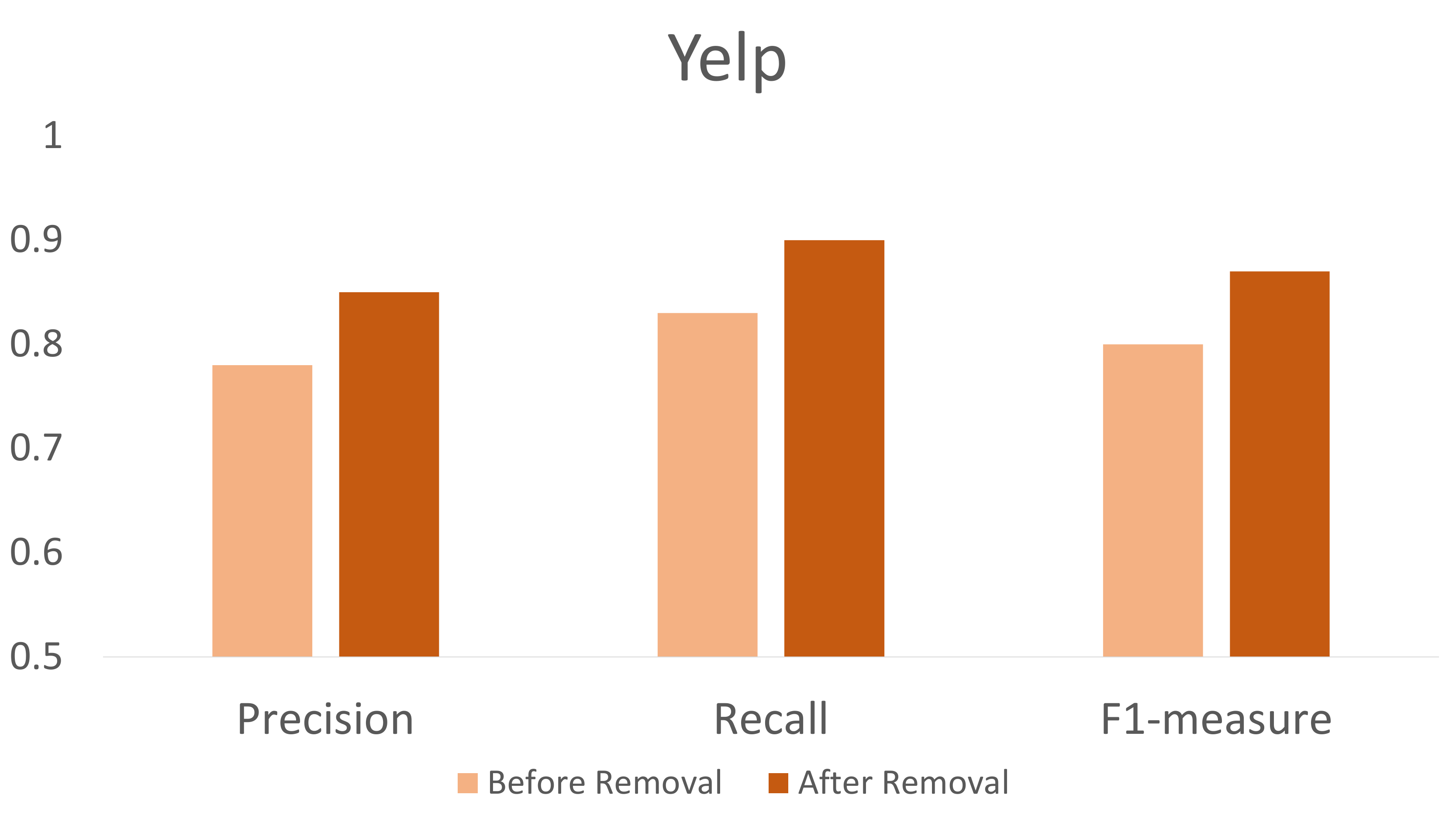}
\caption{The performance of the proposed approach before and after removing deviant reviewers from groups on the Yelp dataset.} 
\label{fig:removal-yelp}
\end{figure}
\begin{figure}[h]
\centering
\includegraphics[width=0.47\textwidth]{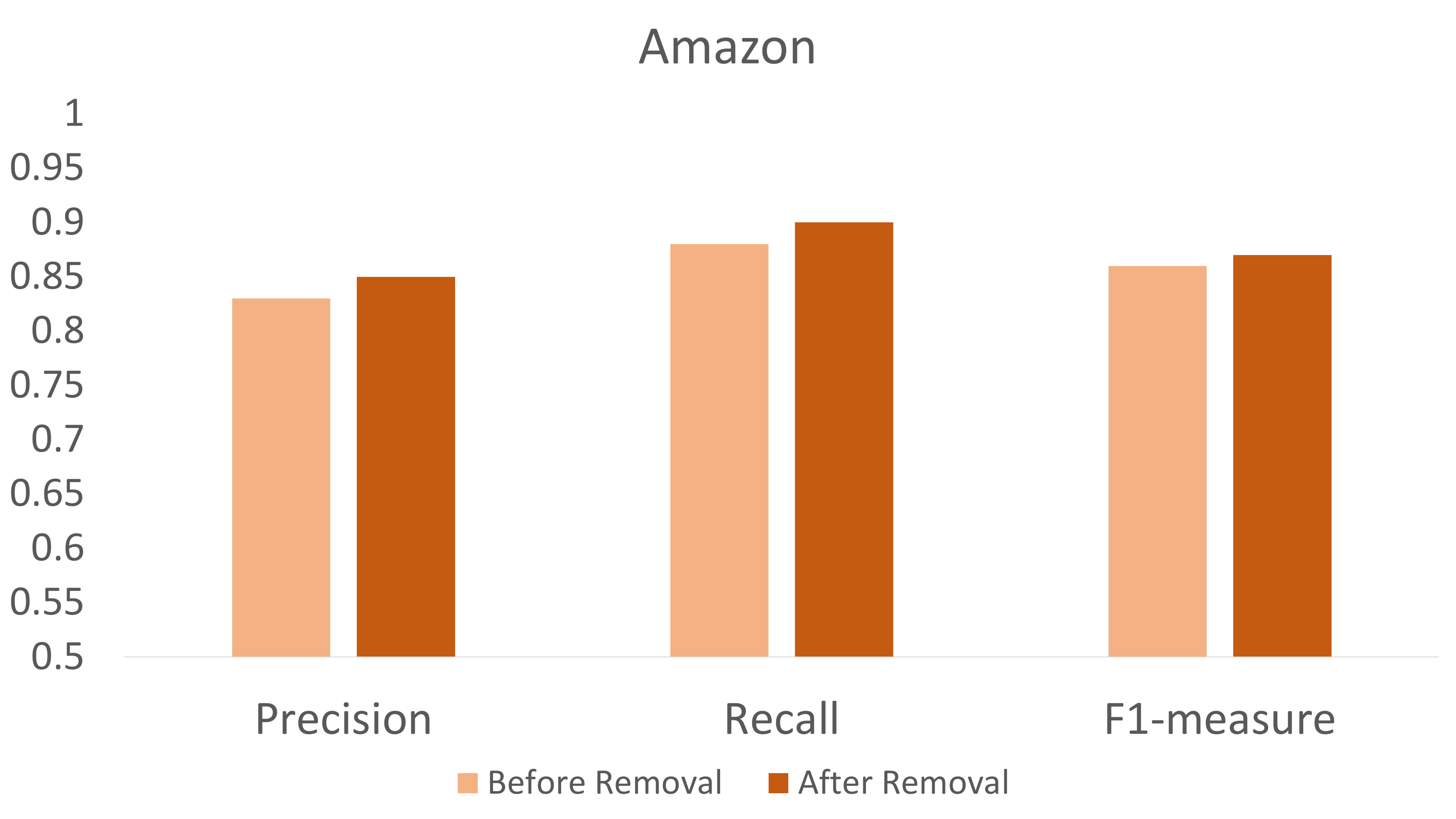}
\caption{The performance of the proposed approach before and after removing deviant reviewers from groups on the Amazon dataset.} 
\label{fig:removal-amazon}
\end{figure}

With such an observation, in the final step (Sec. \ref{sec:classification}) we first removed these deviant reviewers with minimum connections. 
In other words, we removed the reviewers with the least similarity to other reviewers in a group. Figs. \ref{fig:removal-yelp} and \ref{fig:removal-amazon} show the performance of the proposed approach before and after deviant reviewer removal. As a result of removing unintentional genuine reviewers from fraudster groups and also fraudster imposters from a group of genuine reviewers, the FP and FN decrease, respectively. Hence, the precision and recall improve which in turn results in an improvement for F1-value for both Yelp and Amazon datasets.

\section{Conclusion}
\label{sec:conclusion}
Previous studies have investigated the problem of fraudster group detection, but they only rely on handcrafted features for groups, and they are not able to model the non-local semantic dependencies between reviewers in each group. In this study we propose a four step approach to address the challenges of this problem and improve performance: extracting reviewer representation, initializing candidate groups, collaboration modeling using HIN-RNN, and finally removing the deviant reviewers from each group for final classification. The proposed approach outperforms most recent works by 22\%, and 12\% in terms of recall, and F1 on the Yelp dataset, respectively. Future works can use more advanced graph-based networks such as Graph Convolutional Networks (GCNs) to refine the representations of reviewers in each group for final group-level representation.

\bibliographystyle{IEEEtran}
\bibliography{references.bib}

\end{document}